\pdfoutput=1
\documentclass[11pt,a4paper]{article}

\usepackage[margin=25mm]{geometry}
\usepackage[T1]{fontenc}
\usepackage[utf8]{inputenc}
\usepackage{lmodern}
\usepackage{microtype}
\usepackage{graphicx}
\usepackage{pdfpages}
\usepackage{amsmath,amssymb}
\usepackage[numbers,sort&compress]{natbib}
\usepackage[font=small,labelfont=bf]{caption}
\usepackage{booktabs}
\usepackage{xurl}
\usepackage{hyperref}

\hypersetup{
  pdftitle={Landing-induced viscoelastic changes in an anthropomimetic foot joint structure are modulated by foot structure and posture},
  pdfauthor={Satoru Hashimoto, Yinlai Jiang, Hiroshi Yokoi, Shunta Togo}
}

\title{Landing-induced viscoelastic changes in an anthropomimetic foot\\ joint structure are modulated by foot structure and posture}

\author{
Satoru Hashimoto$^{1}$, Yinlai Jiang$^{2}$, Hiroshi Yokoi$^{1,2}$, and Shunta Togo$^{1,2,*}$\\[0.6em]
\small $^{1}$Department of Mechanical and Intelligent System Engineering,\\[-0.1em]
\small Graduate School of Informatics and Engineering,\\[-0.1em]
\small The University of Electro-Communications, Tokyo, Japan\\
\small $^{2}$Center for Neuroscience and Biomedical Engineering,\\[-0.1em]
\small The University of Electro-Communications, Tokyo, Japan\\[0.4em]
\small $^{*}$Corresponding author: \href{mailto:s.togo@uec.ac.jp}{s.togo@uec.ac.jp}
}
\date{}

\begin{document}
\maketitle

\begin{abstract}
How skeletal architecture and landing posture shape the immediate post-impact viscoelastic response of the foot remains incompletely understood, in part because cadaveric specimens are ill-suited to repeated impact testing across postures. In this study, we developed an anthropomimetic foot joint structure aimed at replicating the skeletal geometry of the human foot. Using a vertical drop apparatus that simulates landing and a viscoelastic system-identification model, we investigated how skeletal structure and posture modulate the apparent post-impact viscoelastic response. The results show that the multi-jointed anthropomimetic structure exhibited a higher damping ratio than simplified flat and rigid feet. Moreover, ankle dorsiflexion and toe extension systematically shifted the identified parameters, reducing the damping ratio under the tested conditions. Taken together, these findings indicate that an arch-like, multi-jointed skeletal architecture can enhance impact attenuation in an anthropomimetic mechanical foot, and that morphology and passive posture alone can tune the trade-off between attenuation and rebound. The observed trends are qualitatively consistent with reported differences in human landing strategies, and highlight the engineering advantage of anatomically informed skeletal design for achieving tunable impact attenuation through postural adjustment.
\end{abstract}


\section{Introduction}
Humans acquired bipedal locomotion through a long evolutionary process. During this process, the foot evolved from a structure with opposing grasping functions, as seen in apes, to an arched structure \cite{Susman1983,HarcourtSmith2004,McNutt2018}. It is believed that this structural change in the foot occurred as the living environment of human ancestors changed from trees to the ground \cite{Holowka2018}. When comparing the morphology of the feet of humans and apes, it can be observed that humans have short toes aligned in the same direction, and their metatarsals and tarsals are thick and large \cite{Susman1983,HarcourtSmith2004,McNutt2018}. In addition, they have a well-developed band of elastic tissue called the plantar fascia, which supports the arch of the foot and adaptively changes the rigidity of the foot \cite{Hicks1954}. The human foot is thought to have acquired functions such as shock absorption during landing, adaptation to uneven surfaces, and contribution to propulsion during walking \cite{Ker1987,Ren2008,Venkadesan2020,Welte2021}.

The locomotion styles of terrestrial mammals can be categorized into three types---unguligrade, digitigrade, and plantigrade---based on the morphological characteristics and posture of the foot \cite{Hildebrand1960,Polly2006}. The unguligrade style, observed in animals such as horses and cattle, involves propulsion using hooves. The digitigrade style, seen in animals such as dogs, cats, and birds, involves walking with the heel elevated and making contact with the ground only through the toes. The plantigrade style, found in animals such as bears and rodents, involves placing the entire sole of their foot, including the heel, on the ground. Humans are classified as plantigrades; however, unlike other plantigrades, humans exhibit distinct postural transitions while walking, landing on the heel, and propelling with the toes. Furthermore, during running, humans transition to a digitigrade-like locomotion, landing on the extended toes \cite{Cunningham2010}. Such variations in morphology and posture are expected to alter the balance between energy dissipation and elastic energy storage at foot--ground contact, thereby modulating impact attenuation and rebound. A compact way to quantify this balance is through apparent viscoelastic properties (effective stiffness and damping) identified from the transient force response during landing. Therefore, although various properties of terrestrial mammalian feet have been discussed (e.g., moment arms and resistance to load stress) \cite{Polly2006,Dick2017}, this study specifically focuses on landing-induced viscoelastic responses.

Several simulation analyses using mathematical models have been conducted to investigate the viscoelastic properties of the human foot \cite{Gilchrist1996,Guler1998,Shourijeh2015}. However, the viscoelastic properties identified from such models include contributions from elements such as the soft tissues of the foot (e.g., skin and subcutaneous tissues) and footwear. As a result, the viscoelastic properties inherent to the skeletal and ligamentous architecture itself have not been addressed directly. Furthermore, in some studies, the viscoelastic coefficients of the foot have been determined in relation to posture using mathematical modeling approaches. The findings revealed that the viscous coefficient peaks during foot-ground contact, while the elastic coefficient peaks during toe-off in walking \cite{Takashima2003,Hashimoto2010}. However, the influence of soft tissues has not been excluded in these studies, leaving it unclear whether the foot exhibits viscoelastic changes derived solely from the skeletal and ligamentous architecture. Additionally, because these findings are based on simulations, it is necessary to experimentally investigate the viscoelastic changes in the skeletal and ligamentous architecture of the foot in the real world.

Cadaveric studies have provided valuable data on individual foot structures that are difficult to measure in vivo, such as the elastic properties of the plantar fascia \cite{Wright1964} and the relationship between Achilles tendon tension and plantar fascia strain during toe extension \cite{Carlson2000}. However, this approach has inherent limitations for studying landing dynamics. Postmortem changes and repeated loading can progressively alter tissue properties, making it difficult to obtain consistent measurements across many trials. In particular, investigating posture-dependent viscoelastic changes during landing requires repeated impacts across multiple postures, which is difficult to replicate reliably with biological specimens.

As an alternative to using cadaveric feet, a robotics-based approach can be considered, in which foot mechanics that mimic the structure of the human foot are developed and used for experimentation. This approach enables researchers to adjust various parameters, such as the skeletal structure and mechanical properties of biological tissues, enabling easier analysis of human foot dynamics than using cadaveric feet. Humanoid robotics offer relevant insights from the perspective of engineered feet. Many humanoid robots, such as Honda's ASIMO \cite{Sakagami2002} and Boston Dynamics ATLAS \cite{Guizzo2019}, feature flat-sole foot designs owing to their advantages in terms of control and design simplicity. Several lines of bioinspired foot design have emerged beyond flat soles. Linkage-based mechanisms have been used to approximate the arch structure and its elastic properties during walking \cite{Seo2009,Narioka2012,Hashimoto2013}. Higher anatomical fidelity has been pursued through multi-DOF designs with articulated joints, distributed sensors, or individual toe joints and plantar fascia \cite{Davis2010,Asano2016,Burgess2023}. More recently, adaptive compliance has been emphasized, including deformable soles for uneven terrain, double-arched structures for ground reaction force absorption, and tensegrity-based designs for stability enhancement \cite{Piazza2016,Lee2023,Yeom2024}. Despite these advances, to the best of our knowledge, no prior work has combined precise reproduction of human foot bone geometry based on anthropometric 3D scan data with ligamentous constraints in a single foot structure, nor has the posture-dependent viscoelastic response of such a structure during landing been systematically investigated. The present study addresses both of these gaps by developing an anthropomimetic foot joint structure and using it as a physical testbed to quantify how skeletal structure and landing posture modulate the apparent post-impact viscoelastic response. This approach draws on techniques established in biomimetic robotic hands and arms, where designs replicating human bone and ligament structures have enabled flexible and dexterous mechanisms \cite{Xu2016,Hughes2018,Zhu2023PartI,Zhu2023PartII,Obata2023,Yang2024Shoulder,Yang2024Elbow}.

In this study, we hypothesize that, in an anthropomimetic skeletal-ligamentous architecture, landing posture alone can systematically modulate the apparent viscoelastic response immediately after impact. This study has two objectives: (i) to develop an anthropomimetic foot joint structure aimed at replicating key human skeletal geometry and ligamentous constraints, rather than a generic robotic foot; and (ii) to test this hypothesis experimentally using the developed anthropomimetic foot joint structure as a physical model under controlled landing impacts. The methodology entails the following steps. First, inspired by the techniques used in biomimetic robotic hands \cite{Xu2016,Obata2023}, we fabricated an anthropomimetic foot joint structure using 3D-printed plastic parts and flexible materials. Second, we conducted impact loading experiments by dropping the foot structures while varying their structures, i.e., the shape and degrees of freedom (DoF), and postures, that is, the ankle angle and toe angle. Finally, we analyzed the changes in the viscoelastic properties under different landing conditions by examining the measured impact forces transmitted to the ankle and identifying the viscoelastic coefficients using a simple spring--mass--damper model.

\section{Materials and methods}
\subsection{Experimental design}
\begin{figure}[p]
\centering
\includegraphics[width=0.94\textwidth]{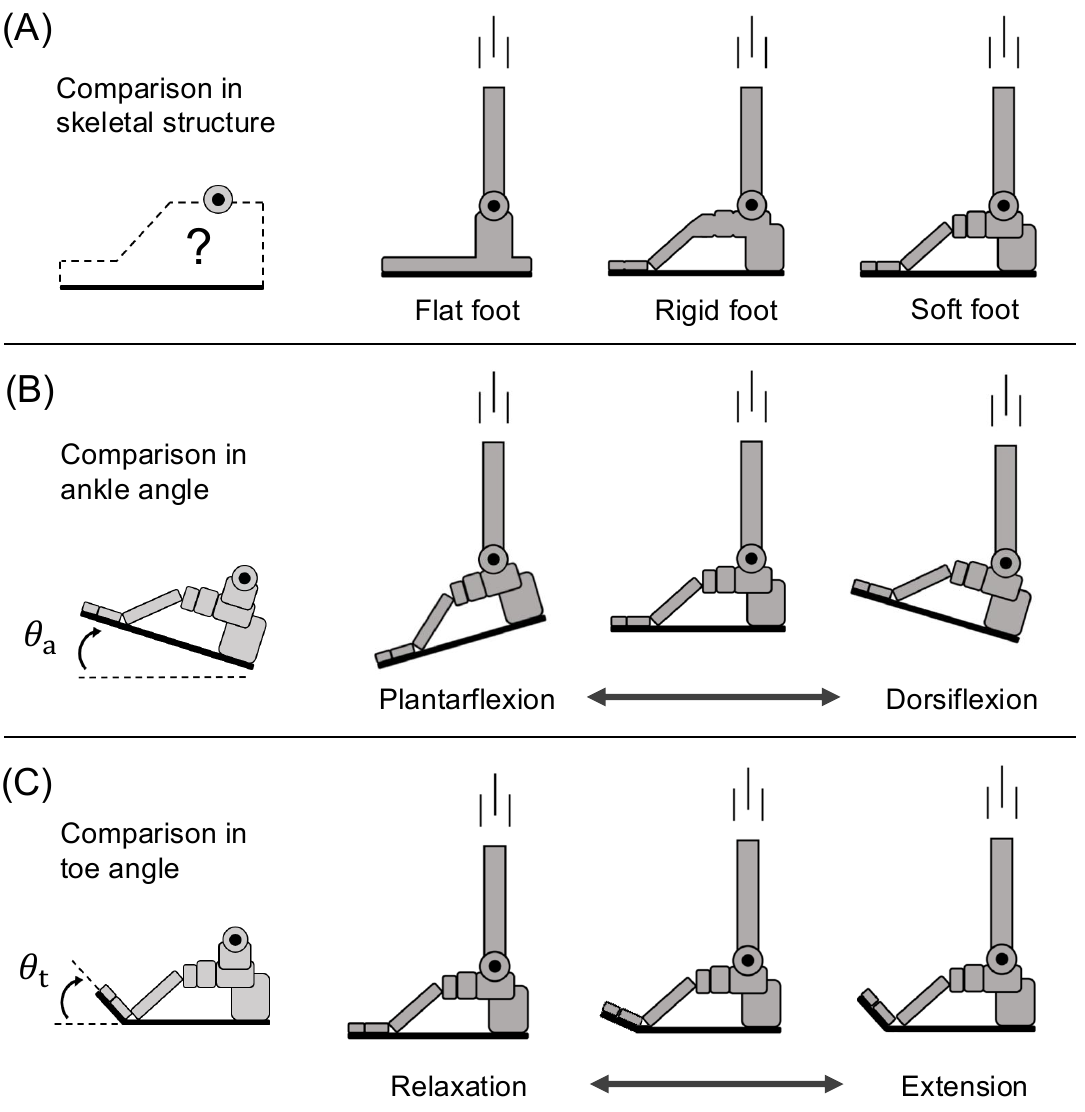}
\caption{Three comparative conditions in the impact loading experiments. (A) Differences in the skeletal structure. The landings of three types of foot structures---flat foot, rigid foot, and soft foot (anthropomimetic foot joint structure)---are compared. (B) Differences in the ankle angle. Landings with varying ankle angles $\theta_a$ are compared. The ankle angle $\theta_a$ is defined as zero degrees when the sole is horizontal, with dorsiflexion considered in the positive direction. (C) Differences in toe angle. Landings with varying toe angles $\theta_t$ are compared. The toe angle $\theta_t$ is defined as zero degrees when the toes are relaxed, with extension considered in the positive direction.}
\label{fig:conditions}
\end{figure}

We developed an anthropomimetic foot joint structure and two simplified counterparts (rigid and flat) with matched overall dimensions and mass, enabling systematic comparisons of skeletal structure and landing posture (Figs.~\ref{fig:overview} and \ref{fig:design}). We then conducted controlled free-fall impact loading experiments to quantify how foot skeletal structure and landing posture modulate the immediate post-landing viscoelastic response. The study comprised three predefined comparisons (Fig.~\ref{fig:conditions}): (1) skeletal structure (flat, rigid, and soft/anthropomimetic), (2) ankle angle $\theta_a$ ($-30^\circ$, $-15^\circ$, $0^\circ$, and $15^\circ$), and (3) toe angle $\theta_t$ ($0^\circ$, $15^\circ$, $30^\circ$, and $45^\circ$) adjusted via the toe-extension tendon. For each condition, the foot was dropped from four initial heights (50, 100, 150, and 200 mm) with 10 repeated trials. Impact force transmitted to the ankle (load cell), foot height change (infrared distance sensor), and landing motion (video) were recorded. We quantified peak force and identified viscoelastic parameters and damping ratio from the impact-force waveform. Details of the foot structures, experimental apparatus, and analysis pipeline are described below.

\subsection{Development of the anthropomimetic foot joint structure}
\begin{figure}[p]
\centering
\includegraphics[width=0.94\textwidth]{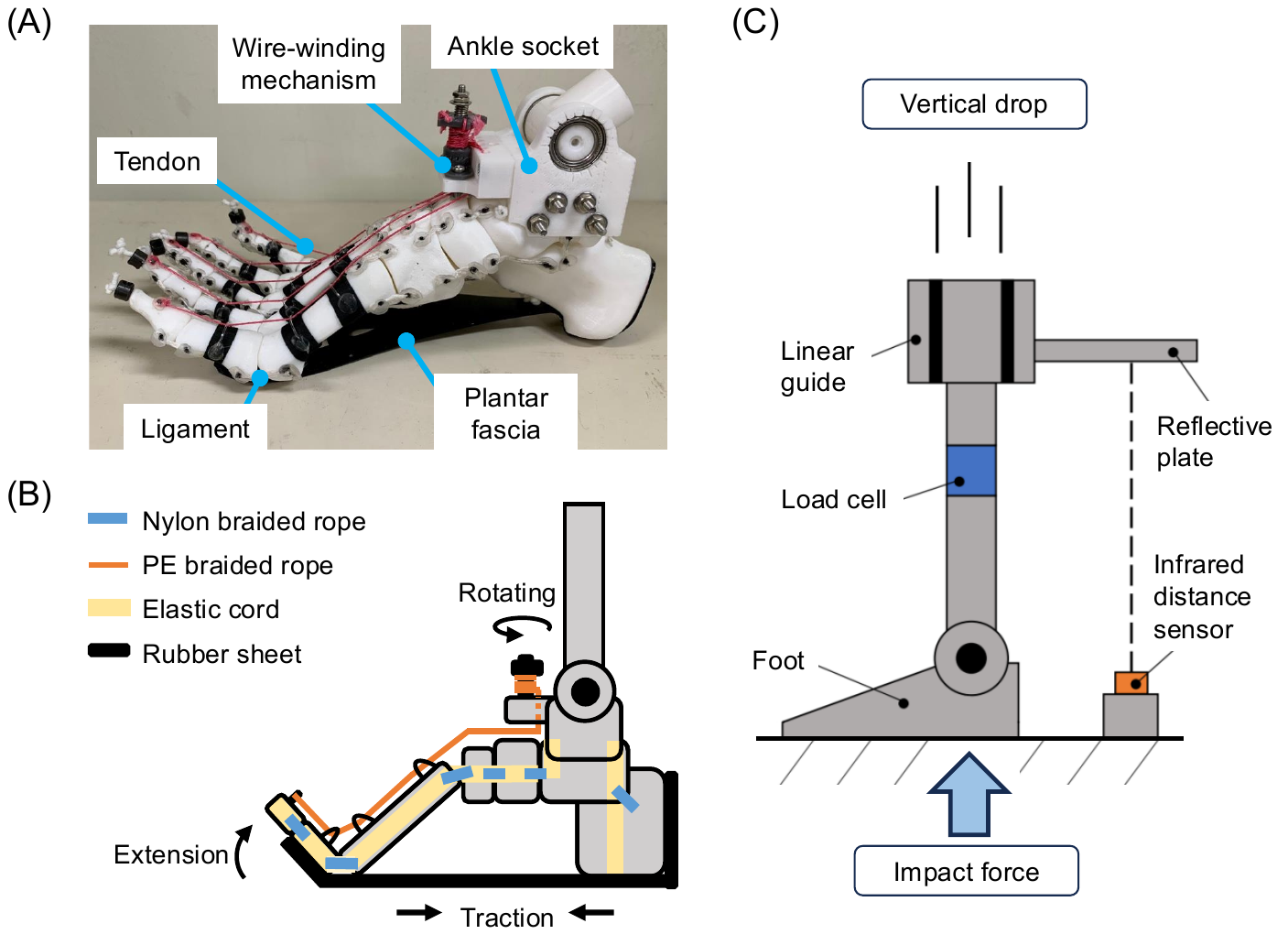}
\caption{Overview of the anthropomimetic foot joint structure and experimental apparatus. (A) Photograph of the fabricated anthropomimetic foot joint structure. (B) Schematic of the anthropomimetic foot joint structure. Nylon braided ropes are attached at positions corresponding to human ligaments to restrict the range of motion. Joint restorative force is incorporated into each joint by threading elastic cords through the hollow, 3D-printed skeletal parts. Polyethylene (PE) braided ropes are used as tendons, enabling toe extension and tensioning of the plantar fascia, represented by a rubber sheet, through a winding mechanism attached to the ankle. (C) Schematic of the experimental apparatus. The fabricated foot joint structure is allowed to free-fall along a linear guide. The impact force transmitted to the ankle is measured using a load cell attached to the ankle, and the position of the foot joint structure is recorded using infrared distance sensors integrated into the experimental apparatus.}
\label{fig:overview}
\end{figure}

\begin{figure}[p]
\centering
\includegraphics[width=0.94\textwidth]{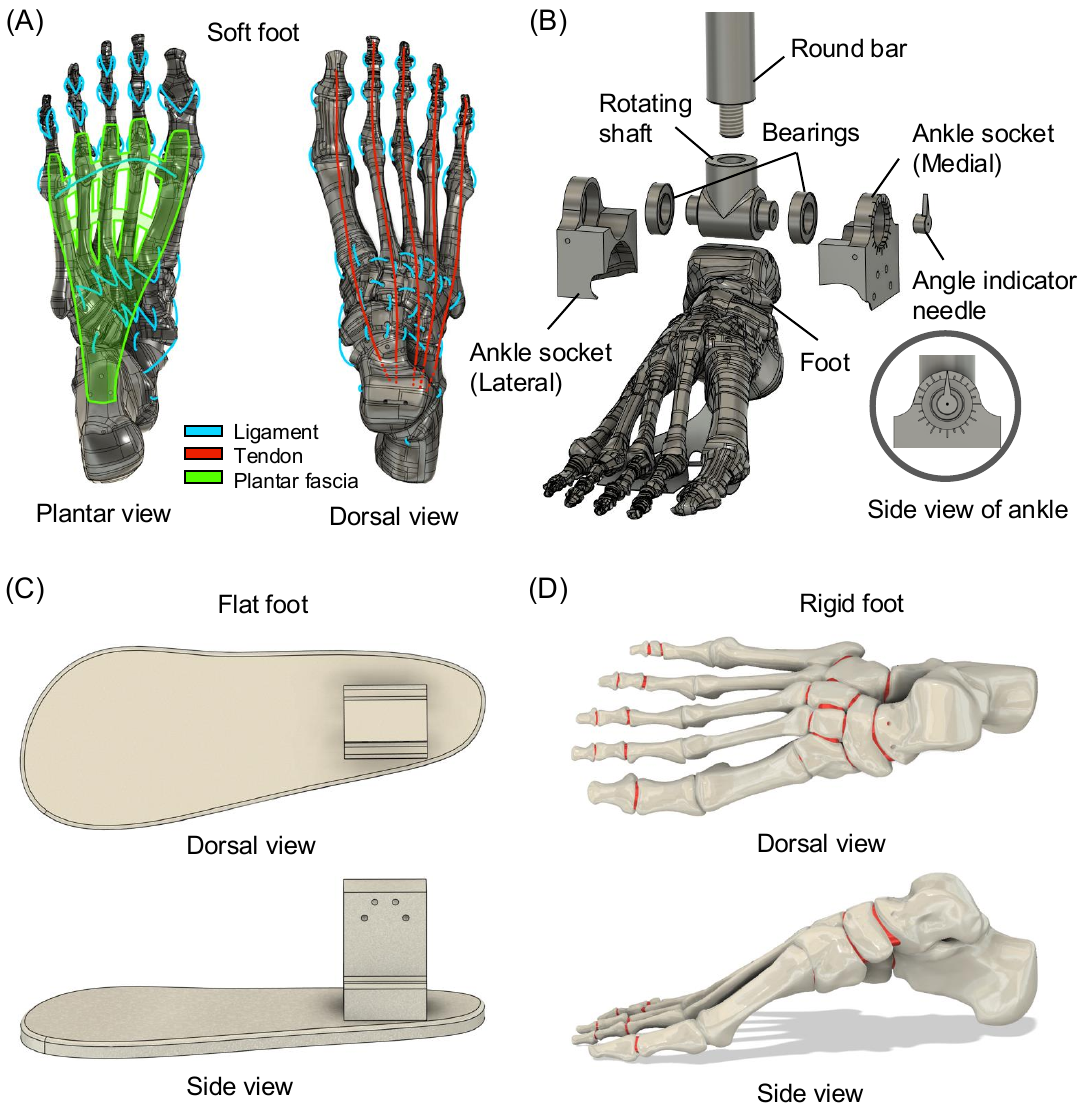}
\caption{Design of each foot structure used in the experiments. (A) Design and attachment positions of the ligaments, tendons, and plantar fascia in the anthropomimetic foot joint structure (soft foot). The attachment positions and shapes of each biological structure are designed based on anatomical illustrations from anatomical references. (B) Design of the ankle socket for adjusting ankle angles. The ankle socket is fixed to the talus using screws and nuts, with a shaft rotating via bearings. The diagram enclosed at the bottom right circle provides a side view of the angle indicator needle used to confirm the ankle angle. (C) 3D CAD model of the flat foot and (D) 3D CAD model of the rigid foot designed for comparative experiments. The red-highlighted areas indicate the regions where the small joints (except for the MTP joint) are rigidly connected.}
\label{fig:design}
\end{figure}

An anthropomimetic foot joint structure (human right foot) designed to approximate key skeletal geometry and ligamentous constraints was developed (Fig.~\ref{fig:overview}). Figure~\ref{fig:design} shows a 3D CAD model (Autodesk Fusion, Autodesk Inc., USA) of the foot and the attachment positions of the ligaments, tendons, and plantar fascia. The 3D CAD data for the developed foot structures (anthropomimetic, rigid, and flat) and the custom 3D-printed parts used in the experimental apparatus are publicly available in a GitHub repository (\url{https://github.com/TogoLab/anthropomimetic-foot-joint-structure}) to facilitate reproducibility. The shapes and attachment positions of each structure were designed based on anatomical references \cite{Schunke2017}.

The skeletal parts of the anthropomimetic foot joint structure were created using 3D scan data obtained from the BodyParts3D web resource provided by the Life Science Integrated Database Center (DBCLS) \cite{BodyParts3D}. The data were edited using the 3D CAD software and 3D printed with a polylactic acid (PLA) filament (Raise3D Pro3, Raise 3D Technologies, Inc., USA). The ligaments that restricted the range of motion of the joints were recreated using braided nylon ropes with a diameter of 1.15 mm, fixed to the joint areas. The ends of the braided ropes were hardened using a glue gun, perforated, and secured using M1 screws. The plantar fascia, which supports the arch structure, was created by cutting a rubber sheet approximately 3 mm thick. The sheet was then screwed onto the calcaneus and proximal phalanges of the toes. To standardize the material of the foot sole during the drop experiments, regardless of the landing posture, the same rubber sheet was attached to the calcaneus. PE-braided ropes with a diameter of 0.52 mm were employed to enable the tendons to extend the toes. These ropes were wound using a mechanism that involved combining a clutch and spring mounted on the ankle (Fig.~\ref{fig:overview}B). The winding mechanism was fabricated using a stereolithography 3D printer (Form 3, Formlabs Inc., USA) with Tough 2000 resin. In human joints, the joint capsule provides a restoring force and maintains the relative positions of bones. The function was replicated in the anthropomimetic foot joint structure by hollowing out the bones and threading elastic cords (diameter of 2 mm) through their interiors.

To adjust the ankle angle of the foot, an ankle socket, as shown in Fig.~\ref{fig:design}B, was 3D printed using PLA filament. The interior of the socket was designed to fit the shape of the talus and secured to the talus using four screws and M3 nuts. In addition, the rotational axis of the ankle, with an attached round rod, was fixed to the ankle socket through the bearings. An angle indicator needle mounted on the rotational axis, combined with grooves marked at $15^\circ$ intervals on the ankle socket, allowed precise measurement of the ankle angle (Fig.~\ref{fig:design}B). The clutch mechanism for winding the toe extension tendon rope was mounted on the ankle socket. The fabricated anthropomimetic foot joint structure had the following dimensions: foot length 218 mm, foot width 82 mm, height (from the sole to the rotational center of the ankle) 88 mm, and weight 136 g.

In addition to the anthropomimetic foot joint structure (soft foot), the flat foot and rigid foot were also fabricated for comparative experiments. The 3D CAD models are shown in Figs.~\ref{fig:design}C and \ref{fig:design}D. The flat foot is a rigid structure with zero DoFs, commonly used in humanoid robots, and features a flat sole. Its external shape was designed based on the 3D bone data used for the anthropomimetic foot joint structure, ensuring that the sole closely follows the contour of the sole of the anthropomimetic foot. The rigid foot was designed by rigidly connecting all the small joints of the anthropomimetic foot joint structure, except for the MTP joint, which resulted in a structure in which only the toes retained mobility (five DoFs). The ligaments of the MTP joint and the internal elastic cord structure for the restoring force were designed in the same manner as the anthropomimetic foot joint structure.

To isolate the effects of skeletal structural differences in the drop experiments, the following parameters were matched across all three foot structures: (i) the same rubber sheet used as the plantar fascia of the anthropomimetic foot was attached to the sole of both the flat foot and the rigid foot, and an identical rubber sheet was also affixed to the calcaneus of the rigid foot; (ii) when fitted with the ankle socket, the height from the sole to the rotational axis was set to 88 mm for all structures; and (iii) oil clay was added near the ankle so that the total weight of each foot matched the anthropomimetic foot weight of 136 g, ensuring equal potential energy at a given drop height across all conditions.

\subsection{Experimental apparatus}
\begin{figure}[p]
\centering
\includegraphics[width=0.94\textwidth]{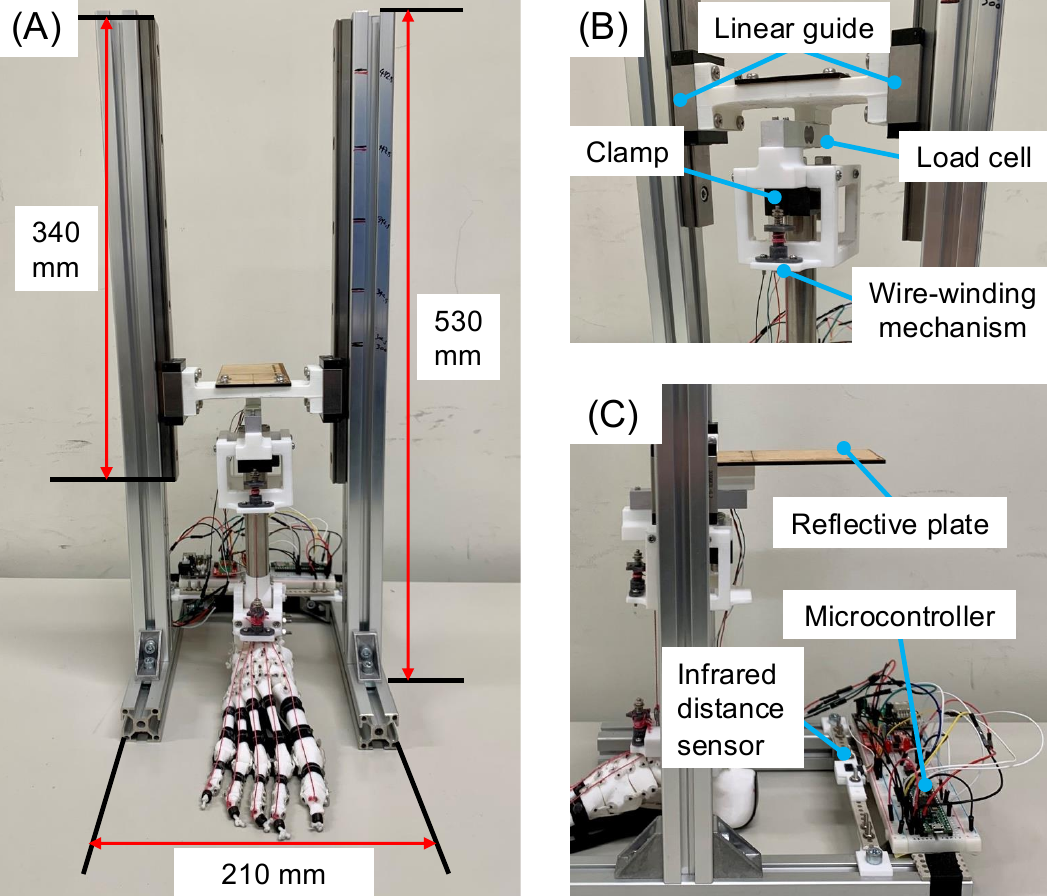}
\caption{Overview of the experimental apparatus. (A) Photograph and dimensions of the experimental apparatus. The apparatus is equipped with linear rails, a load cell, and infrared distance sensors, which allow the foot structures to be dropped vertically while measuring the impact force transmitted to the ankle and foot height. (B) Load cell installation position. The load cell is placed between the two linear guides and the clamp that secures the ankle rod, enabling the forces transmitted to the ankle to be measured. (C) Infrared distance sensor placement. The infrared distance sensors are fixed at the lower part of the experimental apparatus and measure the position of a reflective plate attached for infrared detection. The two sensors are connected to a microcontroller board, and the measured data are recorded on a PC via the microcontroller.}
\label{fig:apparatus}
\end{figure}

An experimental apparatus was constructed to measure the impact force transmitted to the ankle and foot position upon landing after a free fall. Figure~\ref{fig:apparatus}A presents an overview of the experimental apparatus. The main structure of the apparatus was built using $30\,\mathrm{mm}\times30\,\mathrm{mm}$ aluminum frames, and a linear guide and rails were used to implement the vertical drop mechanism for the foot structures. The effects of friction and variations in the drop angle were minimized by allowing the foot structures to slide between the two linear rails, improving the reproducibility of the experiment. To record the behavior of the foot structures during the drop, a camera (iPhone XR, Apple, USA) was fixed in front of the experimental setup.

A single-point load cell (SC133-10 kg, Sensor and Control Co., Ltd, China) was installed to measure the impact force transmitted to the ankle. As shown in Fig.~\ref{fig:apparatus}B, the load cell was fixed between the clamp that secured the round rod and the linear guide using two mounting parts. These two mounting parts were both 3D-printed using a PLA filament. An amplifier (SparkFun Qwiic Scale-NAU7802, SparkFun Electronics, USA) was used to amplify the output voltage of the load cell and perform analog-to-digital conversion. The amplifier communicated with a microcontroller (Teensy 4.1, PJRC, USA) via I2C.

In the impact-loading experiments, an infrared distance sensor (GP2Y0E03, SHARP Corp., Japan) was used to measure the height of the foot from the ground. Figure~\ref{fig:apparatus}C shows the infrared distance sensor mounted on the experimental apparatus and the reflective plate used for the measurement. The sensor has a measurement range of 4--50 cm and can transmit data to a microcontroller via I2C. The sensor was fixed to the frame of the experimental apparatus and a reflective plate was attached to the falling-foot mechanism for measurement. The output values from both the load cell and infrared distance sensor were recorded on a PC via a microcontroller.

A wire traction mechanism was used to adjust the ankle angle during the foot drop. A mechanism similar to the tendon winding system of the anthropomimetic foot joint structure was installed at the top of the round rod, and dorsiflexion and plantar flexion of the ankle were achieved by fixing the end of the wire to the ankle socket (Fig.~\ref{fig:apparatus}B). The PE braided rope, used as the tendon in the anthropomimetic foot joint structure, was also used for this purpose. Additionally, the vertically extended aluminum frame was marked at 50 mm intervals from the ground to allow for adjustment of the drop height during the experiments. The experimental apparatus had the following dimensions: height 530 mm, width 210 mm, depth 260 mm, and linear rail length 340 mm.

\subsection{Experimental procedure and data analysis}
The number of free-fall trials was set to 10 per condition, and the initial drop height $h$ (the height from the ground to the sole) was varied as 50, 100, 150, and 200 mm. The initial drop heights were chosen to span a range from low to moderate impact loading in uniform increments, enabling a systematic parametric investigation of how the apparent viscoelastic properties change with impact magnitude. Because the present setup drops only the foot structure (136 g) without the inertia of the full body, the loading conditions are not intended to replicate physiological ground reaction forces during human locomotion directly. For each trial, the experimenter manually supported the linear guide and released it at designated height marks on the frame of the experimental apparatus, ensuring a controlled free fall from the specified positions.

The sampling rate of each sensor was set to 300 Hz, and the measurement duration was 3 s, with the foot released approximately 1 s after the start of data collection. An offset was applied to the load cell before each trial to ensure that the measured force was 0 N when the foot was unloaded or in contact with the ground. Similarly, for the infrared distance sensor, an offset was applied during post-processing such that the measured height was 0 mm when the foot was in contact with the ground.

In the experiment investigating viscoelastic changes owing to differences in skeletal structure, the experimental apparatus was equipped with the flat foot (Fig.~\ref{fig:design}C), rigid foot (Fig.~\ref{fig:design}D), and soft foot (anthropomimetic foot joint structure; Fig.~\ref{fig:overview}A), each of which was subjected to free fall. The tendons of the soft foot were removed during the experiment to eliminate the influence of factors other than the skeletal structure.

In the experiment investigating viscoelastic changes caused by differences in the ankle angle, the soft foot without tendons was subjected to free fall with the ankle angle $\theta_a=-30^\circ$, $-15^\circ$, and $15^\circ$. For the $\theta_a=0^\circ$ condition, the experimental data from the skeletal structure comparison experiment were used. The ankle angle was adjusted using the angle-indicator needle on the ankle socket and controlled via a wire-winding mechanism attached to the upper part of the ankle. To prevent the winding mechanism from loosening due to vibrations during free fall, the ankle angle was checked and readjusted before each of the 10 trials.

In the experiment investigating viscoelastic changes caused by differences in the toe angle, the soft foot was equipped with the toe extension tendon, and the toe angle $\theta_t=0^\circ$, $15^\circ$, $30^\circ$, and $45^\circ$ for testing. The toe angle was measured using a protractor with the first metatarsal head as the center of rotation and the position of the distal phalanx of the hallux as the reference point. Similar to the ankle angle, the toe angle was checked and adjusted before each of the ten trials to ensure consistency.

The data measured by the load cell and infrared distance sensor were visualized using MATLAB (R2023a, MathWorks Inc., USA). The time-series data from each trial were aligned based on the timing of the peak impact force, and the results were averaged over 10 trials. The measured impact force waveforms (mean $\pm$ standard deviation across 10 trials) for all foot structures, ankle angles, and toe angles at each drop height are provided in the supplementary material (Figs. S2--S5).

\subsection{Simplified model analysis}
\label{sec:model}
Viscoelastic coefficients were identified using a spring--mass--damper model to quantitatively evaluate the viscoelasticity of the foot. In this study, the foot-landing phenomenon was modeled as an impulse response in a simple spring--mass--damper system, where the force transmitted through the spring and damper was considered equivalent to the force measured by the load cell at the ankle. The identification procedure was as follows. First, a spring--mass--damper model with unknown viscoelastic coefficients was prepared. Using the ODE solver in MATLAB Simulink, the system response was computed by applying an impulse input to the model while varying the viscoelastic coefficients. Let $M$ denote the mass of the falling foot, $x$ the displacement of the foot, $k$ the elastic coefficient, $c$ the viscous coefficient, $F$ the impact force generated at the foot sole upon landing, and $F_R$ the force transmitted to the ankle (Fig. S1). The equation of motion for the spring--mass--damper model is given by equation~\eqref{eq:motion}, and force $F_R$ derived from the model is expressed by equation~\eqref{eq:transmitted}:
\begin{align}
M\ddot{x} &= -kx-c\dot{x}+F, \label{eq:motion}\\
F_R &= kx+c\dot{x}. \label{eq:transmitted}
\end{align}
Here, using the drop height of the foot $h$, the gravitational acceleration $g$, and the collision duration during landing $\Delta t$, the force $F$ can be expressed as
\begin{equation}
F=\frac{M\sqrt{2gh}}{\Delta t}.
\label{eq:impulse}
\end{equation}
A schematic of the model analysis and the specific parameter values used in the model analysis are provided in the supplementary material (Fig. S1). All parameters, except for $k$ and $c$, were known and kept identical across all foot conditions. In the simulation, $k$ was varied within the range of $10^3$--$10^7$ N m$^{-1}$, and $c$ was varied within 1--$10^4$ N s m$^{-1}$, and the resulting response waveform of $F_R$ was obtained. To reduce the computational load, $k$ and $c$ were updated using the following equations:
\begin{align}
k &= 10^{(3+4(n-1)/399)}, \label{eq:kgrid}\\
c &= 10^{(4(n-1)/399)}, \label{eq:cgrid}
\end{align}
where $n$ is an integer that satisfies $1\leq n\leq400$. The value of $\Delta t$ was determined as 0.015 s, based on the duration of the main peak observed in the measured impact force waveforms. This value was fixed across all foot structures and posture conditions so that, given the deliberately simplified single-DoF representation, all condition-dependent variation in the foot's response would be captured exclusively by the identified $k$ and $c$, providing a consistent basis for cross-condition comparison. We note that the actual impulse duration may vary somewhat with foot structure and posture; this simplification is discussed further in section~\ref{sec:discussion}. We performed a two-dimensional grid search over $(k,c)$ on a log scale ($400\times400$ combinations) and selected the pair that minimized the weighted waveform error.

Subsequently, the weighted error between the impact force waveforms measured in the experiments and those obtained from the numerical simulations using the model was calculated. The viscoelastic coefficients that yielded the smallest weighted error were considered as the viscoelastic coefficients of the foot used in the experiments. The weights were designed to increase the similarity between the measured and simulated waveforms, particularly focusing on the vibrations that followed the first positive peak of the impact force. Specifically, let $w(s)$ represent the weight at the $s$th sampling point and $s_\text{p}$ represent the time step at which the first positive peak appears in the measured waveform. For $1\leq s\leq s_\text{p}$, the weight was set as $w(s)=1.0$. For $s_\text{p}<s$, let $s_\text{pm}$ denote the time step at which a positive or negative peak appears in the waveform. The weight at these points is defined as follows:
\begin{equation}
w(s_\text{pm})=w(s_\text{pm}-1)+0.05\sqrt{F(s_\text{pm})^2},
\label{eq:weight}
\end{equation}
where $F(s_{pm})$ is the impact force at step $s_\text{pm}$. For all other time steps $s\neq s_\text{pm}$, the weight was set to $w(s)=1.0$. This weighting scheme ensures that the larger and more prolonged the post-first-peak oscillations, the greater the weights assigned to those portions, thereby allowing the identification of viscoelastic parameters that accurately reproduce the vibrational behavior following the first peak.

Viscoelastic coefficient identification was performed for data from each of the 10 trials, and the damping ratio $c/(2\sqrt{Mk})$ was calculated from the identified viscoelastic coefficients. The simulated waveforms obtained from the model analysis for each foot condition are provided in Fig. S2. Finally, the mean values of the elastic and viscous coefficients and damping ratio were calculated for each comparison. To quantify the goodness of fit, the root-mean-square error (RMSE) between the measured and simulated impact force waveforms was calculated for each condition. The RMSE values for all combinations of foot structure, posture, and drop height are reported in the supplementary material (Table S3).

It should be noted that this model relies on several simplifying assumptions: (i) the viscoelastic response is approximated as linear, (ii) the landing impact is treated as an impulse input whose duration $\Delta t$ is fixed at 0.015 s across all conditions, and (iii) the distributed, multi-joint dynamics of the foot structure are represented by a single-DoF lumped-parameter system. Consequently, the identified elastic and viscous coefficients should be interpreted as apparent values that capture the dominant response mode rather than the full complexity of the multi-joint dynamics. The implications of these simplifications are discussed in section~\ref{sec:discussion}.

\subsection{Statistical analysis}
Non-parametric statistical tests and multiple comparisons were conducted using Python to examine the significant differences between the experimental conditions in the peak impact force, identified elastic coefficient, viscous coefficient, and damping ratio. For the results related to differences in the skeletal structure, the Kruskal--Wallis test was first performed to determine whether significant differences existed between groups (significance level $\alpha=0.05$), followed by multiple comparisons using the Steel--Dwass method ($\alpha=0.05$). For the results related to differences in the ankle angle and toe angle, the Friedman test was first conducted to check for significant differences between groups ($\alpha=0.05$), and then Wilcoxon signed-rank tests were performed for each pairwise comparison with the Bonferroni correction applied (adjusted significance level $\alpha=0.008$). The statistical test results are reported in the supplementary material (Tables S1 and S2).

\section{Results}
\subsection{Viscoelastic changes due to differences in foot skeletal structure}
\begin{figure}[p]
\centering
\includegraphics[width=0.95\textwidth]{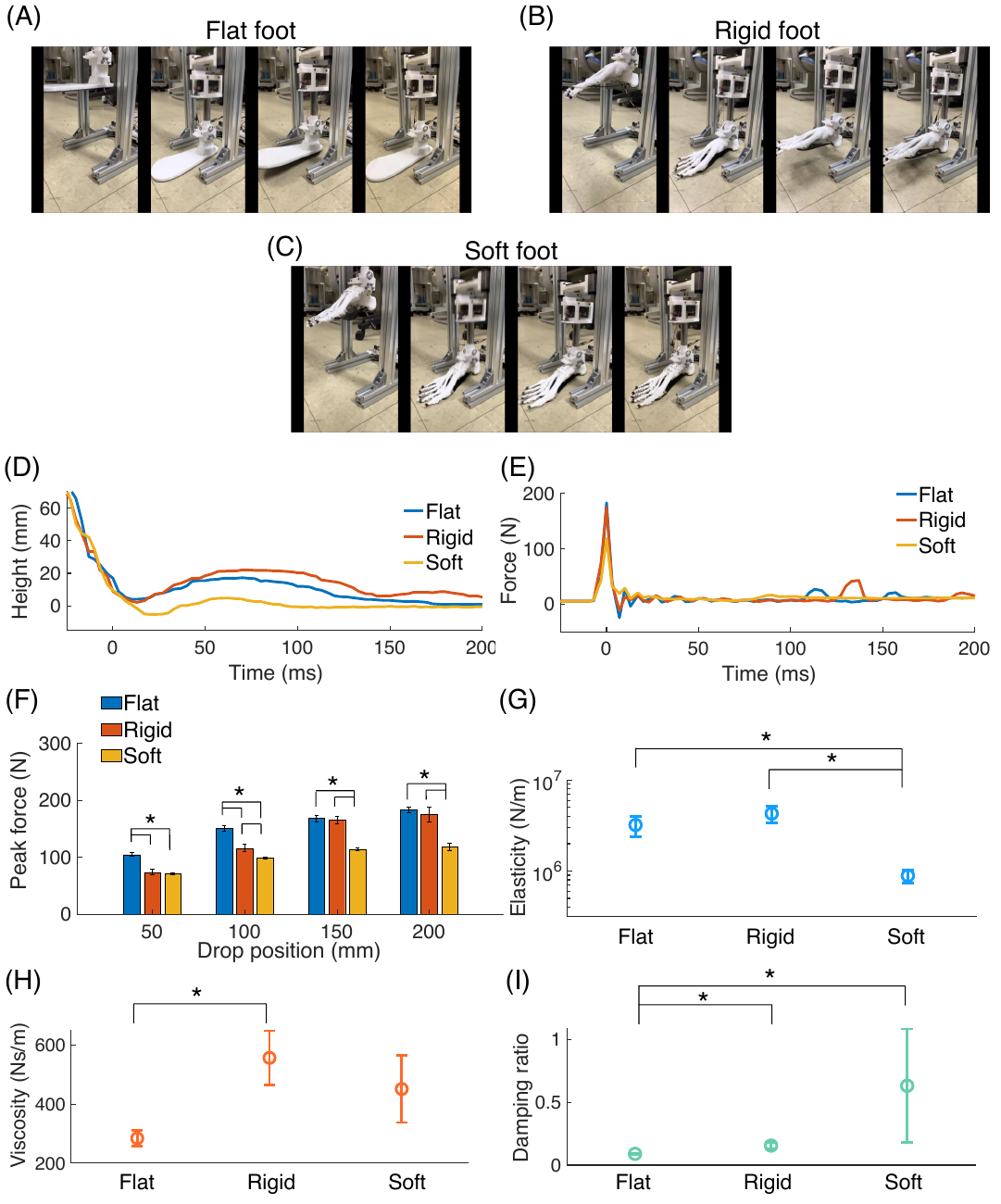}
\caption{Viscoelastic changes caused by differences in the skeletal structure. Snapshots of experiments for (A) flat foot, (B) rigid foot, and (C) soft foot at a drop height of $h=200$ mm. Time-series data of (D) foot height and (E) impact force during landing. Each time-series represents the average of 10 trials, with the moment of impact set as 0 ms. (F) Peak impact forces for each drop height condition for each foot structure. (G) Elastic coefficients, (H) viscous coefficients, and (I) damping ratios estimated using the spring--mass--damper model.}
\label{fig:structure-results}
\end{figure}

Viscoelastic changes associated with differences in the foot skeletal structure are described below (Fig.~\ref{fig:structure-results}). In comparing the behavior of the foot structures at the highest drop height $h=200$ mm, where the impact force was the greatest, the flat foot and rigid-foot mechanisms rebounded from the ground upon landing, while the soft foot structure exhibited compression, with the arch flattening upon impact and subsequently returning to its original shape, achieving stability (Figs.~\ref{fig:structure-results}A--C). The rebound height was approximately 17 mm for the flat foot and 22 mm for the rigid foot, whereas the reduction in the arch height for the soft foot was approximately 5 mm (Fig.~\ref{fig:structure-results}D).

In examining the measured impact force waveforms, the flat foot exhibited the largest peak force, followed by vibrations lasting for approximately 70 ms after the peak (Fig.~\ref{fig:structure-results}E). Similar post-peak vibrations were observed for the rigid foot, although its peak force was smaller and the vibration amplitude was lower than that of the flat foot. The soft foot exhibited the smallest peak force, with no vibrations reaching the negative force range, and the force gradually dissipated over approximately 35 ms. The secondary oscillations observed between 100 and 150 ms for the flat and rigid feet correspond to the impact upon re-contact with the ground following rebound (see Fig.~\ref{fig:structure-results}D), which did not occur for the soft foot owing to its higher damping. A detailed comparison of the peak impact forces showed that the flat foot had relatively higher peak forces at lower drop heights, whereas the rigid foot and soft foot had similarly smaller peak forces (Fig.~\ref{fig:structure-results}F). At $h=50$ mm, statistically significant differences were observed between the flat foot and rigid foot, and between the flat foot and soft foot ($p<0.05$, Table S1), but no significant difference was found between the rigid foot and soft foot. As the drop height increased, the peak force of the rigid foot approached that of the flat foot, whereas the soft foot consistently exhibited the lowest peak force. At $h=200$ mm, significant differences were observed between the flat foot and soft foot, and between the rigid foot and soft foot ($p<0.05$, Table S1), but no significant difference was found between the flat foot and rigid foot.

Because the single-DoF model with a fixed impulse duration is a deliberate simplification (see section~\ref{sec:model}), the identified coefficients reported below should be interpreted as model-dependent apparent descriptors of the prototype response rather than intrinsic mechanical properties. The results of the viscoelastic coefficient identification using the simple spring--mass--damper model revealed that the elastic coefficient was significantly smaller for the soft foot (Fig.~\ref{fig:structure-results}G), while the viscous coefficient was the smallest for the flat foot (Fig.~\ref{fig:structure-results}H). The damping ratio, calculated from the elastic and viscous coefficients, increased in the following order: flat foot, rigid foot, and soft foot (Fig.~\ref{fig:structure-results}I). Statistically significant differences in damping ratios were observed between flat foot and rigid foot, and between the flat foot and soft foot ($p<0.05$, Table S2).

In summary, increasing the DoF in the foot structure, together with the associated ligamentous and elastic-cord compliance at each joint, was associated with greater attenuation under the tested conditions.

\subsection{Viscoelastic changes due to differences in ankle angle}
\begin{figure}[p]
\centering
\includegraphics[width=0.95\textwidth]{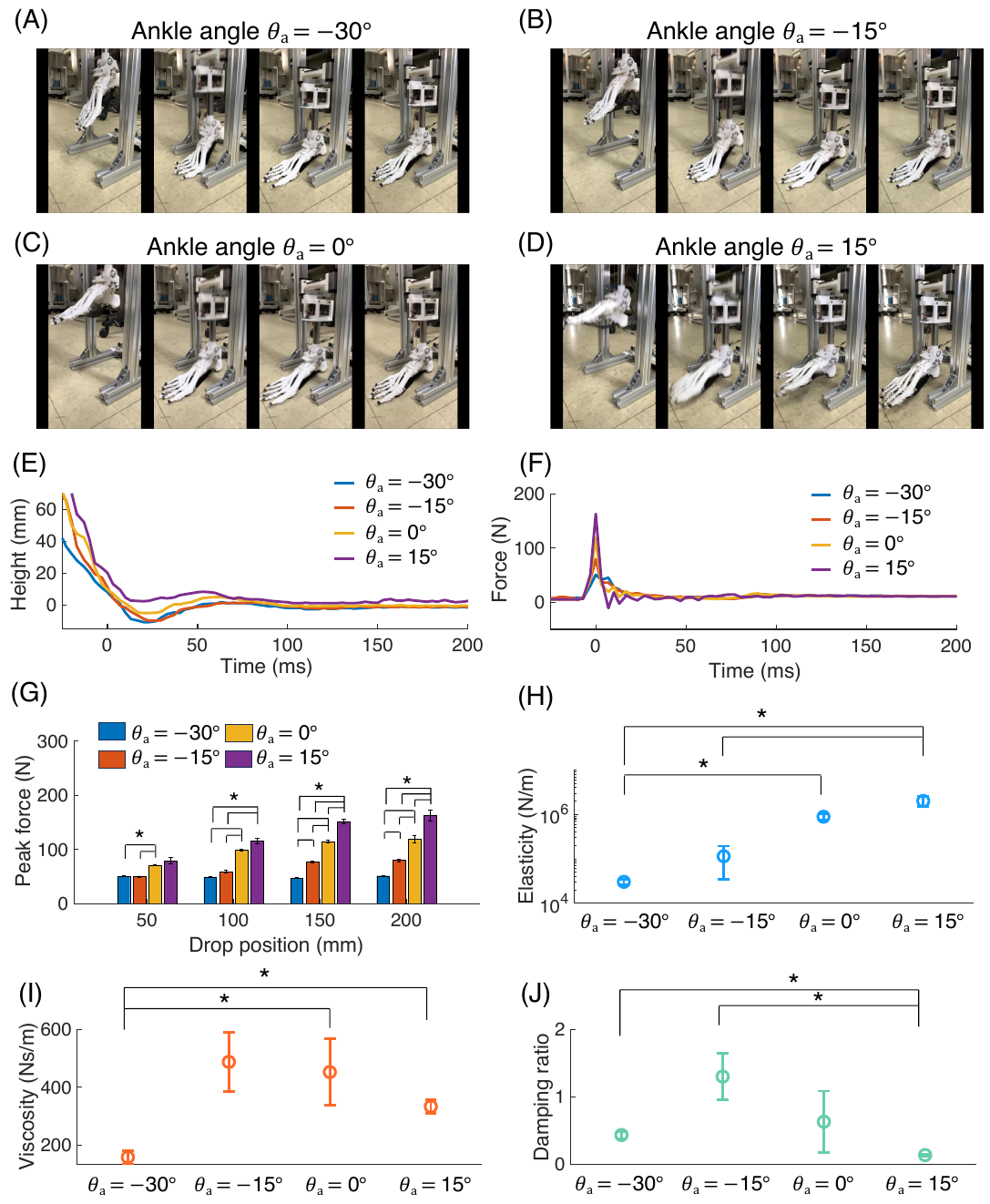}
\caption{Viscoelastic changes caused by differences in the ankle angle in the soft foot (anthropomimetic foot joint structure) without tendons. (A) Snapshots of the experiments at a drop height of $h=200$ mm for $\theta_a=-30^\circ$, (B) $\theta_a=-15^\circ$, (C) $\theta_a=0^\circ$, and (D) $\theta_a=15^\circ$. Time-series data of (E) foot height and (F) impact forces during landing. Each time-series represents the average of 10 trials, with the moment of impact set as 0 ms. (G) Graph summarizing the peak impact forces for each ankle angle across different drop height conditions. (H) Elastic coefficients, (I) viscous coefficients, and (J) damping ratios estimated using the spring--mass--damper model.}
\label{fig:ankle-results}
\end{figure}

The viscoelastic changes associated with variations in ankle angle $\theta_a$ are described below (Fig.~\ref{fig:ankle-results}). The results for $\theta_a=0^\circ$ were consistent with the previously described results for the soft foot. In comparing the foot behavior immediately following landing at a drop height of $h=200$ mm, toe-first landings at $\theta_a=-30^\circ$ and $-15^\circ$ resulted in the ankle rotating in the extension direction upon toe contact, followed by arch flattening when the sole made contact with the ground (Figs.~\ref{fig:ankle-results}A--C). Contrastingly, arch flattening did not occur for heel-first landing at $\theta_a=15^\circ$, and the foot rebounded from the ground after landing (Fig.~\ref{fig:ankle-results}D). The reduction in arch height was approximately 10 mm for $\theta_a=-30^\circ$ and $-15^\circ$, while the rebound height was approximately 7 mm for $\theta_a=15^\circ$ (Fig.~\ref{fig:ankle-results}E). The apparently lower initial height for $\theta_a=-30^\circ$ in Fig.~\ref{fig:ankle-results}E results from the time-alignment procedure: all height data were aligned so that the moment of peak impact force corresponds to 0 ms. Because the $\theta_a=-30^\circ$ condition involves an earlier toe contact followed by a delayed main peak upon full sole contact (see also Fig.~\ref{fig:ankle-results}F), the descent of the foot appears to begin earlier relative to the other conditions in this aligned representation.

In comparing the measured impact force waveforms, $\theta_a=15^\circ$ exhibited the largest peak, accompanied by vibrations lasting approximately 80 ms (Fig.~\ref{fig:ankle-results}F). For $\theta_a=-30^\circ$ and $-15^\circ$, the vibrations dissipated over approximately 40 ms. Notably, for $-30^\circ$, a small initial peak was observed approximately 35 ms before the main peak, which corresponds to toe contact, and the duration of the main peak during sole contact was longer than in other conditions. In examining the peak impact forces across different drop heights, larger $\theta_a$ values (greater dorsiflexion) corresponded to higher peak forces, with the differences becoming more pronounced at higher drop heights (Fig.~\ref{fig:ankle-results}G). At $h=50$ mm, statistically significant differences were observed between $\theta_a=-30^\circ$ and $0^\circ$, as well as $\theta_a=-15^\circ$ and $0^\circ$ ($p<0.008$, Table S1). At $h=200$ mm, all pairwise comparisons of the four ankle angles showed significant differences ($p<0.008$, Table S1). Notably, the peak forces for $\theta_a=-30^\circ$ remained nearly constant regardless of the drop height.

The elastic coefficients identified using the simple spring--mass--damper model were smallest for $\theta_a=-30^\circ$ and increased with larger $\theta_a$ values (Fig.~\ref{fig:ankle-results}H). The viscous coefficients were largest for $\theta_a=-15^\circ$ and tended to decrease with increasing $\theta_a$, with $\theta_a=-30^\circ$ showing a relatively low value (Fig.~\ref{fig:ankle-results}I). Correspondingly, the damping ratios generally decreased as $\theta_a$ increased, although $\theta_a=-30^\circ$ deviated from this trend with an exceptionally low value (Fig.~\ref{fig:ankle-results}J). Significant differences in damping ratios were found between $\theta_a=-30^\circ$ and $15^\circ$, as well as between $\theta_a=-15^\circ$ and $15^\circ$ ($p<0.008$, Table S2).

In summary, toe-first landings tended to show lower peak impact forces and higher damping ratios than heel-first landings within the tested ankle-angle range. However, because the wire traction mechanism used to set the ankle angle allows dorsiflexion but resists plantarflexion, the identified parameters for the flat and heel-first conditions ($\theta_a=0^\circ$, $15^\circ$) may have been influenced by the wire constraint during post-impact rebound, as discussed in section~\ref{sec:discussion}. Accordingly, the ankle-angle-dependent trends reported here should be regarded as qualitative.

\subsection{Viscoelastic changes due to differences in toe angle}
\begin{figure}[p]
\centering
\includegraphics[width=0.95\textwidth]{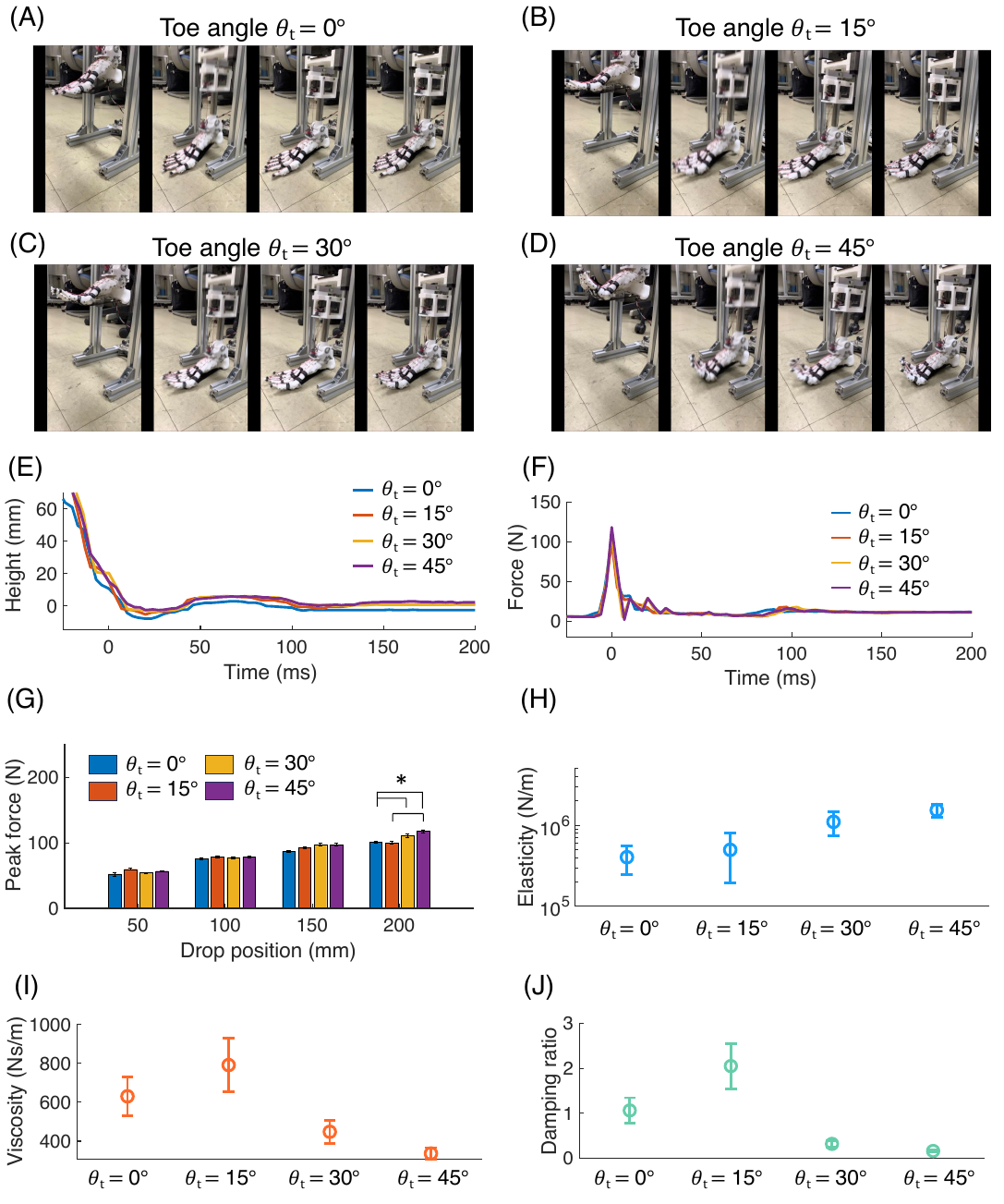}
\caption{Viscoelastic changes caused by differences in the toe angle. (A) Snapshots of experiments at a drop height of $h=200$ mm for $\theta_t=0^\circ$, (B) $\theta_t=15^\circ$, (C) $\theta_t=30^\circ$, and (D) $\theta_t=45^\circ$. Time-series data of (E) foot height and (F) impact forces during landing. Each time-series represents the average of 10 trials, with the moment of impact set as 0 ms. (G) Graph summarizing the peak impact forces for each toe angle across different drop height conditions. (H) Elastic coefficients, (I) viscous coefficients, and (J) damping ratios estimated using the spring--mass--damper model.}
\label{fig:toe-results}
\end{figure}

The results of viscoelastic changes associated with variations in toe angle $\theta_t$ are presented below (Fig.~\ref{fig:toe-results}). The behavior of the foot mechanism while landing at a drop height of $h=200$ mm was consistent across all the toe angles $\theta_t$, with arch flattening occurring upon impact, followed by recovery of the arch shape, and no rebound from the ground (Figs.~\ref{fig:toe-results}A--D). However, a slight trend was observed where the reduction in arch height decreased as $\theta_t$ increased (i.e., as the toes extended further). At $\theta_t=0^\circ$, the arch height reduction was approximately 8 mm, while at $\theta_t=45^\circ$, it was approximately 2 mm (Fig.~\ref{fig:toe-results}E).

In comparing the measured impact force waveforms, no vibrations were observed for $\theta_t=0^\circ$ and $15^\circ$, with the peak force dissipating over approximately 33 ms. By contrast, for $\theta_t=30^\circ$ and $45^\circ$, the peak force was slightly larger, and vibrations persisted for approximately 36 ms at $\theta_t=30^\circ$ and 40 ms at $\theta_t=45^\circ$ after the peak (Fig.~\ref{fig:toe-results}F). The peak impact forces for each drop height condition showed no significant influence of $\theta_t$ at lower drop heights. However, at higher drop heights, increasing $\theta_t$ increased the peak force (Fig.~\ref{fig:toe-results}G). At $h=50$ mm, no significant differences were observed between any of the $\theta_t$ conditions. At $h=200$ mm, significant differences were observed between $\theta_t=0^\circ$ and $30^\circ$, $\theta_t=0^\circ$ and $45^\circ$, and $\theta_t=15^\circ$ and $45^\circ$ ($p<0.008$, Table S1).

Regarding the elastic coefficients identified using the spring--mass--damper model, the smallest value was observed at $\theta_t=0^\circ$, and the elastic coefficients increased with $\theta_t$ (Fig.~\ref{fig:toe-results}H). Conversely, the viscous coefficients tended to decrease as $\theta_t$ increased, with the smallest value observed at $\theta_t=45^\circ$ (Fig.~\ref{fig:toe-results}I). The damping ratios also tended to decrease with increasing $\theta_t$, although the viscous coefficient for $\theta_t=15^\circ$ was unexpectedly high, which resulted in a higher damping ratio for this condition (Fig.~\ref{fig:toe-results}J). No significant differences were observed between any combinations of $\theta_t$ conditions in the results for viscoelastic coefficients or damping ratios.

In summary, greater toe extension tended to increase the identified elastic coefficient and decrease the damping ratio under higher drop heights, consistent with a windlass-like stiffening effect.

\section{Discussion}
\label{sec:discussion}
In this study, we hypothesized that the landing posture in an anatomically grounded anthropomimetic bony and ligamentous foot architecture---designed to replicate key human skeletal geometry and ligamentous constraints rather than a generic robotic foot---can modulate the apparent viscoelastic response observed immediately after impact. To test this hypothesis using a physical testbed, we developed an anthropomimetic foot joint structure and two simplified counterparts (rigid and flat), and conducted controlled free-fall impact loading experiments. Viscoelastic coefficients were identified from the measured impact force waveform using a simplified spring--mass--damper model. The results showed systematic structure- and posture-dependent changes in the identified parameters with (i) the DoF in the skeletal structure, (ii) ankle plantarflexion/dorsiflexion, and (iii) toe extension (Figs.~\ref{fig:structure-results}--\ref{fig:toe-results}). Collectively, these findings indicate that an anthropomimetic multi-jointed skeletal architecture can provide a tunable balance between impact attenuation and rebound in a mechanical foot, and that posture alone can modulate the apparent viscoelastic response in this structure.

The structure-dependent results (Fig.~\ref{fig:structure-results}) suggest that introducing an arch-like, multi-jointed skeletal configuration increases the effective damping ratio during landing, compared with a flat structure. Under vertical loading, the arch deformation (Figs.~\ref{fig:structure-results}C and \ref{fig:structure-results}D) is known to arise from increased joint angles in multiple small joints \cite{Kitaoka1995}. In our physical model, coupling forces across multiple joints likely contribute to the elastic behavior, whereas energy dissipation arises from multiple sources: viscoelastic deformation of the ligament material (nylon braided rope), internal hysteresis of the elastic cords, deformation of the rubber sheet (plantar fascia analogue), and interactions at joint interfaces. The flat and rigid feet, with physically fused joints, exhibited higher resistance to joint displacement, indicating greater apparent elasticity, and this higher elasticity promoted rebound (Figs.~\ref{fig:structure-results}A, \ref{fig:structure-results}B, and \ref{fig:structure-results}G). In contrast, both the rigid and soft feet exhibited higher viscosity than the flat foot, which may be attributed to the dispersion of impact forces across multiple contact regions and the transmission of forces through multiple joints and the associated energy dissipation within the compliant elements connecting them (Fig.~\ref{fig:structure-results}H). For the soft foot in particular, the greater number of articulated joints increases the total deformation experienced by ligaments and elastic cords during impact, which may lead to viscosity dominating over elasticity; consequently, peak impact force decreased and the damping ratio increased (Figs.~\ref{fig:structure-results}F and \ref{fig:structure-results}I). However, we note that the present experiments do not allow us to isolate the individual contributions of each dissipation mechanism, and this decomposition remains a task for future work. These mechanical interpretations are consistent with the notion that multi-jointed, arch-like architectures can enhance impact attenuation.

The damping ratio $c/(2\sqrt{Mk})$ exhibited greater trial-to-trial variability than either the elastic or viscous coefficient alone (Figs.~\ref{fig:structure-results}G--I). As shown in Figs.~\ref{fig:structure-results}G and \ref{fig:structure-results}H, the elastic coefficient exhibited comparable variability across the three foot structures, whereas the viscous coefficient was notably more variable for the rigid and soft feet than for the flat foot. Because the flat foot has zero DoF and no sliding joint interfaces, its energy-dissipation behavior is expected to be highly reproducible. In contrast, the rigid foot (five DoFs at the MTP joint) and the soft foot (full multi-joint articulation) contain compliant joint interfaces where subtle trial-to-trial differences in joint sliding, ligament engagement, and arch deformation may alter the effective viscous dissipation. This joint-related variability in the viscous coefficient may then be amplified in the damping ratio through two pathways: the direct dependence of the damping ratio on $c$, and, for the soft foot in particular, the substantially smaller elastic coefficient (Fig.~\ref{fig:structure-results}G), which reduces the denominator $2\sqrt{Mk}$ and could magnify the effect of any variation in $c$ on the damping ratio. We therefore consider that the elevated damping-ratio variability likely reflects a combination of inherent mechanical variability at compliant joint interfaces and the sensitivity of the damping-ratio formula when the elastic coefficient is small. This observation also highlights a representational limit of the single-DoF spring--mass--damper model, which lumps distributed multi-joint dynamics into two scalar parameters.

The ankle angle results (Fig.~\ref{fig:ankle-results}) demonstrate that the apparent viscoelastic response of the developed anthropomimetic foot joint structure can be tuned by landing posture at the ankle. In our experiments, heel-first landings exhibited higher apparent elasticity (Fig.~\ref{fig:ankle-results}H), whereas toe-first landings exhibited higher apparent viscosity (Fig.~\ref{fig:ankle-results}I). In the mechanical structure, a heel-first posture likely transmits impact forces to the ankle through fewer joints, thereby reducing the opportunity for energy dissipation within compliant elements and increasing rebound. Conversely, a toe-first posture transmits forces through multiple joints, engaging more ligaments and elastic elements and thus increasing energy dissipation, resulting in a higher damping ratio (Fig.~\ref{fig:ankle-results}J). Qualitatively, these posture-dependent trends are consistent with reported differences in human landing strategies (e.g., heel-first contacts in economical walking versus softer toe-involved contacts in certain landing tasks) \cite{Bramble2004,McKinley1992,Zhang2000}. In particular, Zhang et al. \cite{Zhang2000} reported that the energy absorption distribution across lower-extremity joints changes with landing technique (soft versus stiff), indicating that posture-related modulation of impact response is a well-documented phenomenon in human biomechanics. The present results suggest that part of such posture-dependent modulation may originate from the passive mechanics of the foot skeleton itself, independently of active muscular control. In the present physical testbed, active neuromuscular control and proximal joints are not represented; therefore, the comparison to human strategies is qualitative.

It should also be noted that the wire traction mechanism introduces a directional confound. For the flat and heel-first conditions ($\theta_a=0^\circ$, $15^\circ$), the wire may have partially resisted plantarflexion during post-impact rebound, effectively adding an external restoring force that could elevate the apparent elastic coefficient and suppress the amplitude of post-peak oscillations. Because no wire-free control experiments were conducted, the magnitude of this effect cannot be quantified; however, the overall trend of increasing apparent elasticity with greater dorsiflexion (Fig.~\ref{fig:ankle-results}H) would be expected to persist even if the wire contributed to the absolute values at $\theta_a=0^\circ$ and $15^\circ$, because the toe-first conditions ($\theta_a=-30^\circ$, $-15^\circ$) involved dorsiflexion upon landing, which slackens the wire and renders its influence negligible. Therefore, the ankle-angle-dependent results are best interpreted as qualitative trends rather than precise quantitative comparisons across all four angles. Notably, the simulation results for $\theta_a=-30^\circ$ indicated an anomalously low viscous coefficient (Fig.~\ref{fig:ankle-results}I), suggesting that the simple spring--mass--damper model can incur substantial modeling errors under certain conditions (Fig. S2F). This motivates future work with more detailed models.

The toe angle results (Fig.~\ref{fig:toe-results}) indicate that toe extension can alter the apparent viscoelastic parameters under higher landing loads in the developed structure, in line with biomechanical accounts of the windlass mechanism \cite{Hicks1954,Welte2021}. In our structure, increasing toe angle under high-load conditions reduced the damping ratio (Figs.~\ref{fig:toe-results}G and \ref{fig:toe-results}J). A plausible mechanical explanation is that increased tension in the plantar-fascia-like element raises contact pressure across small joints, increasing coupling forces (enhancing elasticity) while reducing sliding at joint interfaces (reducing viscous resistance) (Figs.~\ref{fig:toe-results}H and \ref{fig:toe-results}I). These trends are consistent with prior analyses of foot viscoelastic models \cite{Takashima2003,Hashimoto2010}. We interpret these results as evidence that toe posture, through a windlass-like stiffening effect, can shift the trade-off between rebound and dissipation in an anthropomimetic foot joint structure; however, quantitative equivalence to human in vivo behavior remains to be established. It should be noted that, in the toe angle experiments, the toe extension tendon was attached to enable controlled toe extension, meaning that the system configuration differed slightly from the tendon-free condition used in the ankle angle experiments, even at $\theta_t=0^\circ$. Therefore, the toe angle results should be interpreted as relative comparisons across $\theta_t$ values within this specific configuration, rather than as direct comparisons with the ankle angle experimental conditions.

From an engineering perspective, the present results corroborate the findings of prior bioinspired foot designs. For example, Davis and Caldwell \cite{Davis2010} demonstrated shock absorption and energy storage in a fully articulated humanoid foot, and Piazza et al. \cite{Piazza2016} showed that intrinsic adaptivity of an arch structure can coexist with rigid stance support in the SoftFoot. Our results confirm that these beneficial properties of multi-jointed arch structures also hold in a foot whose skeletal geometry faithfully reproduces that of human bones. In addition, the finding that the apparent viscoelastic response can be adjusted by posture alone suggests that foot posture may serve as an additional passive control parameter for tuning landing behavior in legged robots.

In the broader context of biomimetic robotics, an important contribution of the present work is that the developed anthropomimetic foot joint structure can serve as a physical testbed for dynamic loading scenarios, such as landing impacts during walking and running, that are difficult to investigate experimentally in human subjects. Cadaveric feet are limited in their capacity for repeated testing and systematic posture variation, and computational models depend on assumptions about material and contact conditions. In contrast, the present prototype enables repeatable experiments under controlled conditions while allowing the effects of skeletal structure and posture to be examined independently.

However, this testbed has several limitations. The present design prioritized geometric fidelity of the skeleton and the topological arrangement of joints, and the remaining elements involve the following constraints. First, the prototype does not include soft tissues such as skin, subcutaneous tissue, or the calcaneal fat pad, all of which contribute to impact absorption in the human foot. Future work could incorporate soft-tissue layers using silicone or gel materials. Second, the materials used for the ligaments (nylon braided rope), the joint restoring elements (elastic cords), and the plantar fascia analogue (rubber sheet) were selected for practical fabricability and functional analogy rather than for quantitative correspondence to human tissue mechanics; their mechanical properties have not been characterized and are not expected to match those of their biological counterparts. Therefore, the identified coefficients should be interpreted as apparent properties of the present engineered skeletal-ligamentous prototype, rather than as quantitative estimates of the human foot. In particular, the nylon braided ropes, elastic cords, and rubber sheet likely contributed substantially to the observed stiffness and damping by constraining joint motion, restoring joint posture, and dissipating energy during impact. Measuring the mechanical properties of these materials and performing a quantitative comparison with human tissue data is an important task for future work. Third, the total mass of the prototype (136 g) is lighter than the human foot skeleton, and because the 3D-printed PLA parts have a different density from human bone, the mass distribution also differs. The use of filling materials or alternative 3D-printing materials could improve mass and mass distribution fidelity. Fourth, in the ankle angle experiments, a wire traction mechanism was used to set the ankle angle. This wire allows dorsiflexion but resists plantarflexion. For the toe-first conditions ($\theta_a=-30^\circ$, $-15^\circ$), the ankle rotates toward dorsiflexion upon landing (Figs.~\ref{fig:ankle-results}A and \ref{fig:ankle-results}B), so the wire slackens and its influence is expected to be minimal. However, for the flat and heel-first conditions ($\theta_a=0^\circ$, $15^\circ$), the wire may have partially resisted plantarflexion during post-impact rebound (Figs.~\ref{fig:ankle-results}C and \ref{fig:ankle-results}D), and therefore the force oscillation characteristics at these angles may be partially influenced by the wire constraint. Future work should employ a wire-free fixation mechanism, such as a mechanical lock, to eliminate this confound. Fifth, the present setup does not include knee or hip joints, nor active muscle-tendon actuation; consequently, the observed viscoelastic responses reflect only the passive mechanics of the isolated foot structure. Integration into a multi-joint legged platform would enable evaluation under more dynamic locomotion tasks. Sixth, the single-DoF spring--mass--damper model with a fixed impulse duration is a deliberate simplification, and in vivo validation of the observed trends has not been performed. Introducing higher-order models and comparing the mechanical trends with human experimental data remain tasks for future work.

Beyond robotics, these findings may be relevant to several applied fields. In sports injury prevention, understanding how the passive mechanical properties of the foot skeleton contribute to impact distribution may inform risk assessment for stress fractures and other overuse injuries. In rehabilitation and footwear design, the observation that arch compliance and toe posture modulate impact attenuation characteristics could provide guidelines for optimizing plantar pressure distribution and insole design. In addition, since abnormal plantar pressure distribution is a known risk factor for diabetic foot ulceration \cite{Caselli2002}, insights into how skeletal architecture passively redistributes impact loads may provide a mechanical perspective that could inform future studies on ulcer-preventive orthotic interventions.

\section{Conclusion}
In this study, we developed an anatomically grounded anthropomimetic foot joint structure aimed at replicating key human skeletal geometry and ligamentous constraints, and used it as a physical testbed to examine how morphology and posture shape the immediate post-landing viscoelastic response. Using controlled free-fall impact loading experiments and spring--mass--damper identification from impact-force waveforms, we demonstrated systematic structure- and posture-dependent modulation of apparent viscoelastic parameters. The identified parameters are model-dependent apparent descriptors tied to the present experimental assumptions; nevertheless, the systematic trends across conditions support the following conclusions. Compared with a flat foot and a rigid foot, the multi-jointed anthropomimetic structure exhibited greater impact-force attenuation and a higher damping ratio during landing, indicating that increased skeletal DoF can enhance energy dissipation in this mechanically realistic architecture. Ankle posture qualitatively tuned the landing response, with toe-first landings tending to reduce peak impact forces and increase damping relative to heel-first landings, and toe extension under higher drop heights tending to increase the effective elastic coefficient and reduce the damping ratio, consistent with a windlass-like stiffening effect in the developed structure. Collectively, these results show that morphology and passive posture alone can tune the trade-off between impact attenuation and rebound in an anthropomimetic mechanical foot, with potential implications for the design of robotic feet and for informing future studies in sports biomechanics, rehabilitation, and footwear design. Future work will focus on improving anatomical and material fidelity and integrating the developed structure into a legged robotic platform with knee and hip joints.

\section*{Acknowledgment}
This work was supported by the Japan Society for the Promotion of Science (JSPS) KAKENHI under Grant JP22K04025, Grant JP23H00166, and Grant JP26K00910, and in part by JKA through Promotion Funds from KEIRIN RACE.

\section*{Data availability statement}
The data that support the findings of this study are openly available in the project repository \cite{Hashimoto2026Data}.

\section*{Conflict of interest}
The authors declare that this research was conducted in the absence of any commercial, financial, or non-financial interests.

\section*{Author contributions}
\noindent\textbf{Satoru Hashimoto:} Conceptualization (lead), Data curation (lead), Formal analysis (lead), Investigation (lead), Methodology (lead), Resources (lead), Visualization (lead), Writing -- original draft (lead).\\
\textbf{Yinlai Jiang:} Supervision (supporting), Validation (supporting), Writing -- review \& editing (supporting).\\
\textbf{Hiroshi Yokoi:} Supervision (supporting), Validation (supporting), Writing -- review \& editing (supporting).\\
\textbf{Shunta Togo:} Conceptualization (equal), Funding acquisition (lead), Investigation (equal), Project administration (lead), Resources (equal), Supervision (equal), Validation (equal), Visualization (equal), Writing -- review \& editing (lead).

\clearpage
\includepdf[pages=-,pagecommand={\thispagestyle{empty}},fitpaper=true]{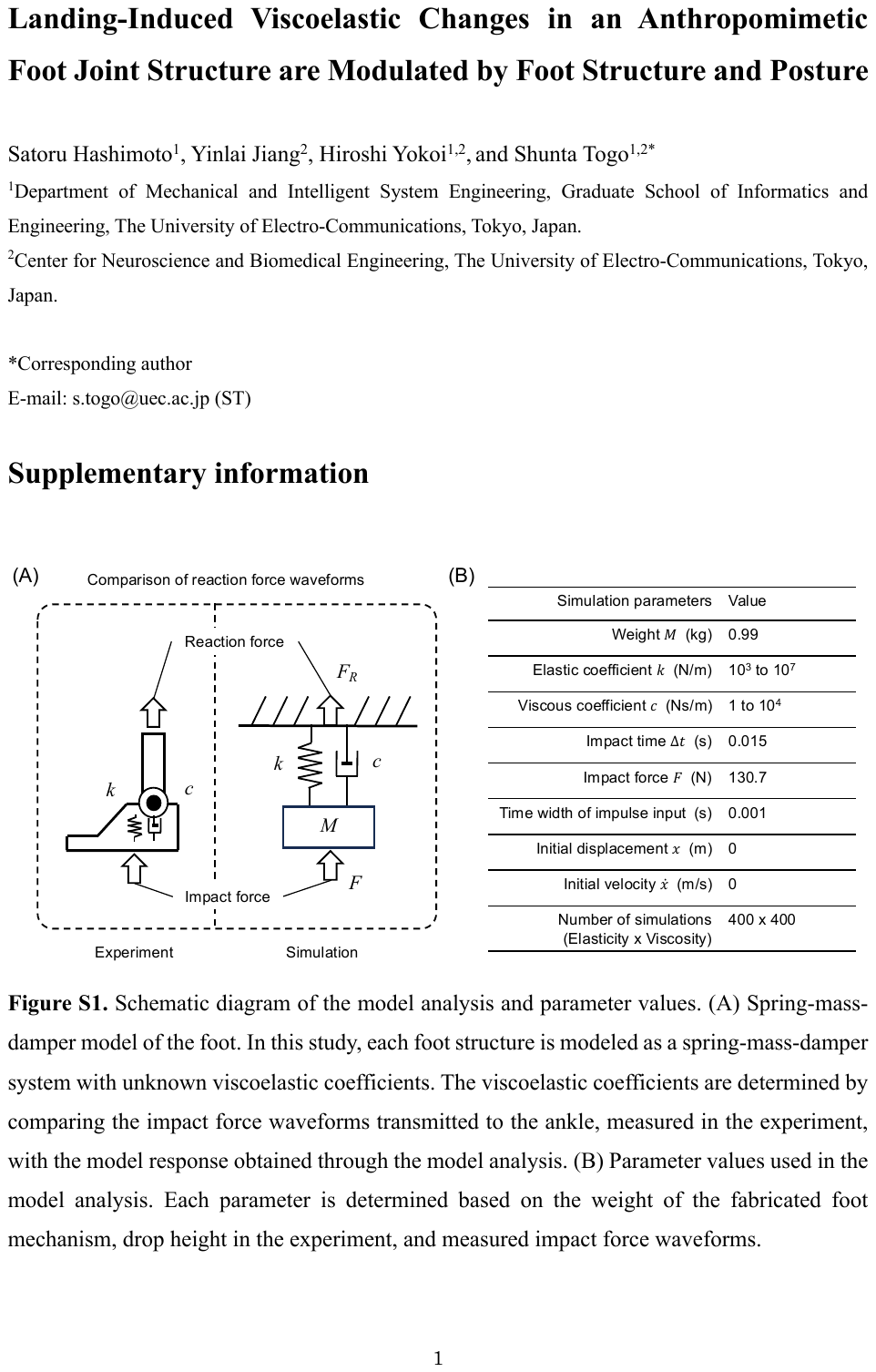}

\end{document}